\documentclass[10pt,twocolumn,letterpaper]{article}

\usepackage[pagenumbers]{cvpr} 

\usepackage{placeins}
\definecolor{cvprblue}{rgb}{0.21,0.49,0.74}
\usepackage[pagebackref,breaklinks,colorlinks,allcolors=cvprblue]{hyperref}

\title{InsightEdit: Towards Better Instruction Following for Image Editing}
\author{
    Yingjing Xu\textsuperscript{1,2}\thanks{Co-first authors.} , Jie Kong\textsuperscript{2}\footnotemark[1] \thanks{Primary Corresponding author.} , Jiazhi Wang\textsuperscript{2}, 
    Xiao Pan\textsuperscript{1}, Bo Lin\textsuperscript{1}\thanks{Co-Corresponding author.} , Qiang Liu\textsuperscript{2} \\
    \\
    $^1$Zhejiang University,  
    $^2$01.ai \\
    {\tt\small \{poppyxu, xiaopan, rainbowlin\}@zju.edu.cn, \{kongjie, wangjiazhi, liuqiang\}@01.ai}
}

\usepackage{graphicx}    
\usepackage{float}
\usepackage{booktabs}
\usepackage{amssymb} 
\usepackage{xcolor}
\usepackage{hyperref}  

\usepackage{pifont}  
\newcommand{\cmark}{\ding{51}}
\newcommand{\xmark}{\ding{55}}

\begin{document}
\maketitle

\begin{abstract}
In this paper, we focus on the task of instruction-based image editing. Previous works like InstructPix2Pix, InstructDiffusion, and SmartEdit have explored end-to-end editing.
However, two limitations still remain: First, existing datasets suffer from low resolution, poor background consistency, and overly simplistic instructions. Second, current approaches mainly condition on the text while the rich image information is underexplored, therefore inferior in complex instruction following and maintaining background consistency. Targeting these issues, we first curated the AdvancedEdit dataset using a novel data construction pipeline, formulating a large-scale dataset with high visual quality, complex instructions, and good background consistency. 
Then, to further inject the rich image information, we introduce a two-stream bridging mechanism utilizing both the textual and visual features reasoned by the powerful Multimodal Large Language Models (MLLM) to guide the image editing process more precisely. 
Extensive results demonstrate that our approach, InsightEdit, achieves state-of-the-art performance, excelling in complex instruction following and maintaining high background consistency with the original image. 
The project page is \href{https://poppyxu.github.io/InsightEdit_web/}{https://poppyxu.github.io/InsightEdit\_web/}.

\end{abstract}   
\begin{figure*}[!ht]
    \centering
    \includegraphics[width=\textwidth]{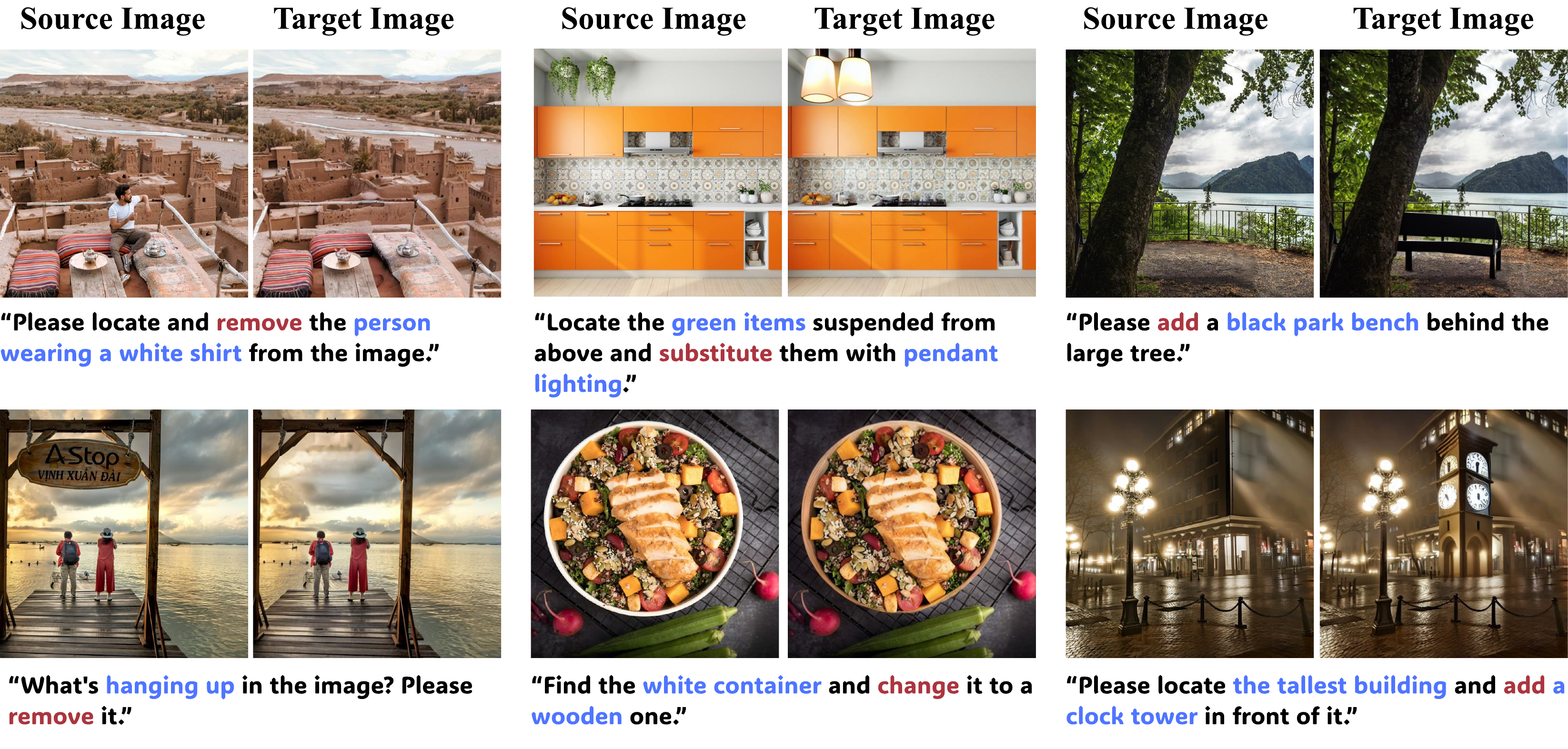}  
    \caption{We propose \textbf{InsightEdit}, an end-to-end instruction-based image editing model, trained on high-quality data and designed to fully harness the capabilities of Multimodal Large Language Models (MLLM), achieving high-quality edits with strong instruction-following and background consistency.}
    \label{fig:showcase}
\end{figure*}
\section{Introduction}
Image editing involves altering an image's appearance, structure, or content, encompassing a range of changes from subtle adjustments to major transformations~\cite{huang2024diffusion}. It has witnessed significant advancements in image editing tasks alongside the development of diffusion models~\cite{wang2023imagen, brooks2023instructpix2pix, cao2023masactrl, geng2024instructdiffusion, ju2024brushnet}, leading to impressive results. 

Existing works like InstructPix2Pix~\cite{brooks2023instructpix2pix}, MagicBrush~\cite{zhang2024magicbrush}, and InstructDiffusion~\cite{geng2024instructdiffusion} have investigated end-to-end image editing. Some recent efforts~\cite{huang2024smartedit, fu2023guiding} explore using MLLM to comprehend instructions, thereby enhancing the ability to cope with complex editing tasks. However, two challenges still remain: (i) \textbf{Lack of high-quality datasets.} As illustrated in Table \ref{tab:data_comparison}, current datasets suffer from low resolution, poor visual quality and background consistency, and overly simplistic, template-based instructions. Specifically, most datasets rely on Prompt2Prompt~\cite{hertz2022prompt} method, which struggles to achieve precise control over the generated images and fails to maintain consistency in unedited regions. These limitations hinder the image editing model's capability in both complex instruction following and high-fidelity target image generation. 
(ii) \textbf{Lack of rich image condition.}
Current methods primarily use the CLIP text encoder to provide conditions, yet it often exhibits limited ability to understand the instructions. Some approaches further leverage the more advanced MLLM to better comprehend instructions. However, they still mainly focus on understanding instructions at the textual level, while neglecting to capture the rich visual semantics of the image, therefore showing weak performance in complex instruction following and maintaining background consistency.

\begin{table*}[t]
\setlength{\abovecaptionskip}{0cm}  
\centering
\small
\setlength\tabcolsep{3.0pt}
\begin{tabular}{@{} l c c c c c c @{}}
\toprule
\textbf{Dataset}            & \textbf{\begin{tabular}[c]{@{}c@{}}Editing \\ Pairs\end{tabular}} & \textbf{Resolution} & \textbf{\begin{tabular}[c]{@{}c@{}}Good Background \\ Consistency\end{tabular}} & \textbf{\begin{tabular}[c]{@{}c@{}}Complex \\ Instructions\end{tabular}} & \textbf{\begin{tabular}[c]{@{}c@{}}High Visual\\ Quality\end{tabular}} & \textbf{\begin{tabular}[c]{@{}c@{}}Automated \\ Pipeline\end{tabular}} \\ \midrule
{InstructPix2Pix}~\cite{brooks2023instructpix2pix}    & 313,010      & $512^2$       & \textcolor{red}{\xmark}                    & \textcolor{red}{\xmark}              & \textcolor{red}{\xmark}        & \textcolor{green}{\cmark}  \\
{MagicBrush}~\cite{zhang2024magicbrush}         & 10,388       & $1024^2$      & \textcolor{green}{\cmark}                    & \textcolor{red}{\xmark}              & \textcolor{green}{\cmark}     & \textcolor{red}{\xmark}    \\
{HIVE}~\cite{zhang2024hive}               & 1,100,000+   & $512^2$      & \textcolor{red}{\xmark}                    & \textcolor{red}{\xmark}              & \textcolor{red}{\xmark}        & \textcolor{green}{\cmark}  \\
{FaithfulEdit}~\cite{chakrabarty2023learning}       & 52,208       & $1024^2$      & \textcolor{red}{\xmark}                    & \textcolor{red}{\xmark}              & \textcolor{green}{\cmark}     & \textcolor{green}{\cmark}  \\
{UltraEdit}~\cite{zhao2024ultraedit}          & 4,108,262    & $512^2$     & \textcolor{green}{\cmark}                  & \textcolor{green}{\cmark}            & \textcolor{red}{\xmark}        & \textcolor{green}{\cmark}  \\
{HQEdit}~\cite{hui2024hq}             & 187,350      & $900^2$       & \textcolor{red}{\xmark}                    & \textcolor{red}{\xmark}              & \textcolor{green}{\cmark}     & \textcolor{green}{\cmark}  \\
{EditWorld}~\cite{yang2024editworld}          & 10,000+      & $512^2$       & \textcolor{red}{\xmark}                    & \textcolor{green}{\cmark}            & \textcolor{green}{\cmark}     & \textcolor{green}{\cmark}  \\ \hline
\textbf{AdvancedEdit (Ours) }     & \textbf{2,536,674}    & $\mathbf{1024^2}$        & \textcolor{green}{\cmark}                  & \textcolor{green}{\cmark}            & \textcolor{green}{\cmark}     & \textcolor{green}{\cmark}  \\ \hline
\end{tabular}
\caption{\textbf{Comparison between previous datasets and ours.} We provide more detailed comparisons in the appendix. }
\label{tab:data_comparison}
\end{table*}

To address the dataset issue, we propose an automated data construction pipeline to generate editing pairs with complex instructions and strong background consistency. Leveraging the perceptive capabilities of MLLM, our approach extracts detailed object information from high-resolution images and utilizes an advanced mask-based editing model to produce realistic and controllable edits. Using this method, we introduce AdvancedEdit Dataset, a comprehensive collection comprising over 2,500,000 editing pairs with high visual quality, complex instructions, and good background consistency. 


To address the lack of rich image condition,  we employ a two-stream bridging mechanism to integrate both high-level textual and rich visual information into the denoising process of the diffusion model. This interaction effectively extracts the conditional information, enabling precise and controllable image editing. 

Our method achieves state-of-the-art performance on both the Reason-Edit~\cite{huang2024smartedit} and AdvancedEdit-Eval, demonstrating its strong capability in complex instruction following and maintaining background consistency.

In summary, our contributions are as follows:
\begin{enumerate}
    \item We propose an automated data construction pipeline for image editing, specifically engineered to facilitate the training of models for complex instruction-based image editing tasks.
    \item  We produce the AdvancedEdit Dataset, a large-scale, high-quality, and fine-grained image editing dataset with complex instructions and good background consistency. Additionally, we introduce the AdvancedEdit-Eval dataset to assess the model’s ability to handle complex instructions.
    \item We present InsightEdit, which utilizes a two-stream bridging mechanism to guide the image editing process with both textual and image features reasoned by MLLM. Extensive experiments are conducted to prove the model's ability to follow complex instructions and maintain high background consistency. 
\end{enumerate}

\section{Related Work}
\subsection{Image Editing Methods}
Recent advancements in image editing can be broadly categorized into mask-based and mask-free approaches. Mask-based image editing~\cite{wang2023imagen, avrahami2023blended, xie2023smartbrush, xie2023dreaminpainter} requires the original image, instructions, and mask image indicating the regions to be edited as input. The mask image typically needs to be manually specified by the user. Blended Latent Diffusion~\cite{avrahami2023blended} extends Latent Diffusion Model~\cite{rombach2022high} by integrating them into a localized image editing framework, enabling the targeted redrawing of specific regions during the denoising process. 
BrushNet~\cite{ju2024brushnet} is inspired by the architecture of ControlNet~\cite{zhang2023adding}. It employs a dual UNet~\cite{ronneberger2015u} structure, where an additional branch is used to enhance feature extraction from the masked regions. Power-Paint~\cite{zhuang2023task} focuses on the distinct characteristics of various image editing tasks by learning a token to classify different editing tasks. Mask-based approaches enable localized redrawing and fine-grained control, resulting in a stable and high-visual-quality output. However, they require additional mask information and are sensitive to mask shape and the specific editing task.

Mask-free methods~\cite{brooks2023instructpix2pix, geng2024instructdiffusion, huang2024smartedit, koh2024generating} present a promising alternative by reducing the dependency on explicit masks and allowing for more flexible and intuitive image editing. InstructPix2Pix~\cite{brooks2023instructpix2pix} utilizes GPT-3~\cite{mann2020language} and a Prompt2Prompt~\cite{hertz2022prompt} methodology to construct an image editing dataset. InstructDiffusion~\cite{geng2024instructdiffusion} builds upon the network design of InstructPix2Pix, aiming to unify various vision tasks through joint training. MGIE~\cite{koh2024generating} enhances instruction-based image editing by learning to generate expressive instructions with MLLM. SmartEdit~\cite{huang2024smartedit} utilizes MLLM to generate text embeddings, leveraging the MLLM's advanced reasoning and comprehension capabilities.

Although instruction-based image editing methods do not require explicit mask guidance, making them more intuitive and convenient, they often yield lower image quality than mask-based approaches. They also face challenges in following complex instructions and maintaining background consistency.

\subsection{Image Editing Datasets}
\label{sec:dataset}
Several works have recently introduced image editing datasets. InstructPix2Pix~\cite{brooks2023instructpix2pix} introduces a large-scale image editing dataset created through fine-tuned GPT-3 and a Prompt-to-Prompt methodology integrated with Stable Diffusion. MagicBrush~\cite{zhang2024magicbrush} further provides a limited-scale, manually annotated dataset for instruction-guided real image editing. Subsequent work, HIVE~\cite{zhang2024hive} introduces more training triplets and human ranking results to provide stronger supervisory signals for improved model training. FaithfulEdits~\cite{chakrabarty2023learning} employed inpainting techniques followed by a filtering process using Visual Question Answering (VQA) models. UltraEdit~\cite{zhao2024ultraedit} explores region-based data generation, employing mask generation to facilitate object redrawing in specified areas. Some works explore the diversity of instructions, such as EditWorld~\cite{yang2024editworld}, which defines and categorizes instructions grounded in various world scenarios. 

As shown in Table~\ref{tab:data_comparison}, despite the notable contributions of these datasets, current image editing datasets still face several challenges, including the inability to scale the data construction process, low dataset resolution, poor background consistency, and a lack of fine-grained instructions for complex understanding and reasoning tasks.
\section{Dataset Construction}

\begin{figure*}[ht]    
    \centering    
    \setlength{\abovecaptionskip}{0cm}
    \includegraphics[width=1.0\linewidth]{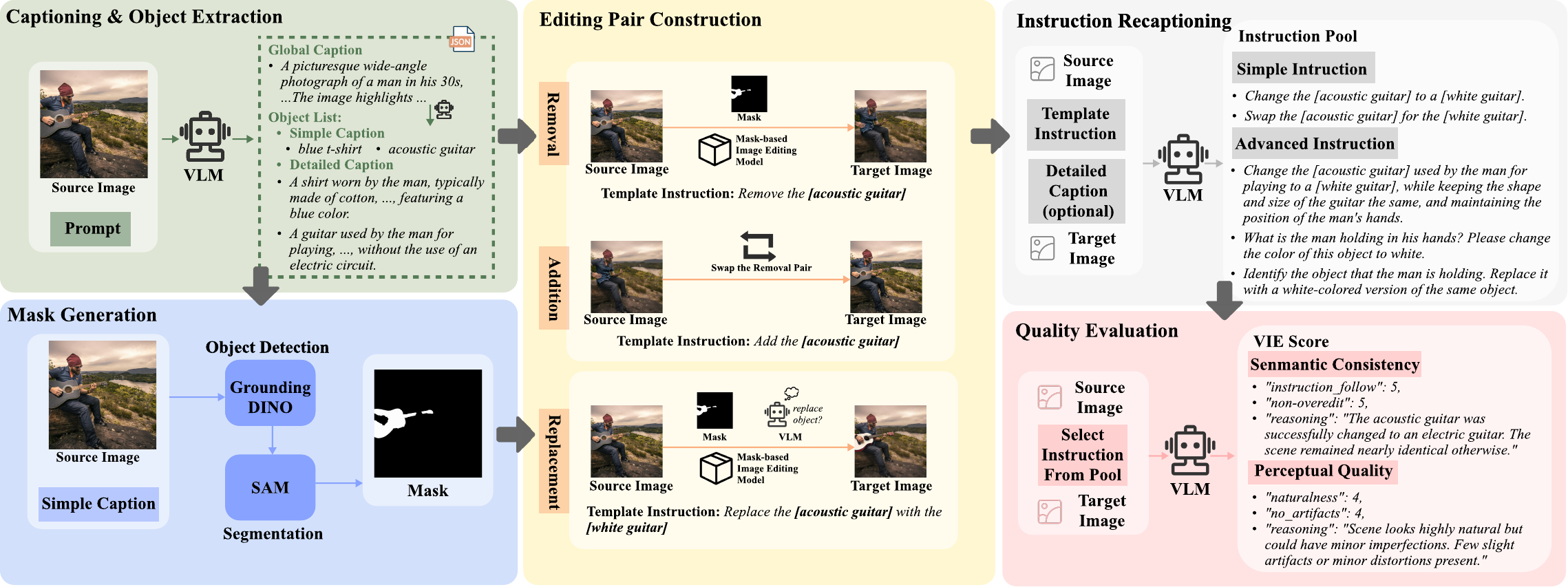}    
    \caption{\textbf{The overall data construction pipeline.}
    \textbf{(1) Captioning \& Object Extraction:} Utilizing VLM to generate a global caption from the source image, and further get an object JSON list contains both simple caption and detailed caption. 
    \textbf{(2) Mask Generation:} Utilizing GroundedSAM to obtain the corresponding mask of each object. 
    \textbf{(3) Editing Pair Construction:} Utilizing mask-based image editing model to construct target image and templated instruction. 
    \textbf{(4) Instruction Recaptioning:} Utilizing VLM to rewrite instruction to gain diverse instructions. 
    \textbf{(5) Quality Evaluation:} Filtering the datasets using VIEScore.}  
    \label{fig:data_construction}    
\end{figure*}


\subsection{Automated Pipeline}
\label{sec:caption_mask_preparation}

We propose an automated data construction pipeline focused on generating high-fidelity, fine-grained image-editing pairs with detailed instructions that demonstrate advanced reasoning and understanding. We categorize the image editing tasks into three types: removal, addition, and replacement. Figure \ref{fig:data_construction} presents our data preparation workflow.

\noindent\textbf{Step 1: Caption \& Object Extraction.}
We leverage the advanced comprehension capabilities of MLLM to generate a global caption that effectively conveys the image's content. Using this caption, we utilize LLM to extract a JSON list of objects, identifying those with physical significance. Each object is defined by a simple caption (e.g., \textit{blue T-shirt}) and a detailed description (e.g., \textit{A shirt worn by the man, typically made of cotton, featuring a blue color}).

\noindent\textbf{Step 2: Mask Generation.} For mask generation, we apply GroundedSAM~\cite{ren2024grounded} to extract local masks for each object, filtering out low-confidence masks based on a predefined threshold, thus obtaining accurate object and mask pairs for further processing.


\noindent\textbf{Step 3: Editing Pair Construction.} As mentioned in Section \ref{sec:dataset}, mask-based image editing models exhibit superior image generation capabilities, offering better control over the generation process for specific tasks. In the process of constructing the edited images, we employed the state-of-the-art mask-based methods~\cite{ju2024brushnet, zhuang2023task} to generate the target image, enabling the creation of more fine-grained and controllable image editing pairs. We categorized the image editing generation tasks into three types: removal, addition, and replacement.

For removal, we apply a mask-based approach, using both the original image and the object mask to generate the post-removal image, simultaneously generating and storing the template instruction: \textit{``remove the [object].''} For addition, we swap the source and target images. The template instruction is: \textit{``add the [object].''}

For the replacement task, we leverage MLLM's creative capabilities to propose an alternative object for the scene, often resulting in plausible and innovative outcomes. Simultaneously, we generate and store the template instruction: \textit{``replace the [source object] with the [target object].''}.

\noindent\textbf{Step 4: Instruction Recaptioning.}
To diversify the complexity of the instructions, we recaption the instructions into both simple and advanced versions. The simple instructions are generated through synonym replacement and task template modifications. For instance, the instruction \textit{``replace the [acoustic guitar] with [white guitar]''} is rewritten as \textit{``Swap the [acoustic guitar] for the [white guitar].''}

In contrast, the advanced instructions are designed to increase complexity in two key ways. First, we replace simple object descriptions with detailed ones prepared in Step 1 and apply different task templates. Second, we recaption instructions to gain reasoning capabilities, such as \textit{``What is the man holding in his hands? Please change the color of this object to white.''} This approach challenges the editing model to generate images with more intricate details and greater precision. 

\noindent\textbf{Step 5: Quality Evaluation.}
To construct a high-quality dataset, we assess and filter the image-editing pairs using VIEScore~\cite{ku2023viescore}, which aligns evaluations with human preferences via MLLM. The evaluation framework consists of two components: semantic consistency, which measures instruction following and evaluates whether the image has been over-edited, and perceptual quality, which evaluates image fidelity, including the presence of artifacts or blurriness. The rating example  of VIEScore can be seen in Figure \ref{fig:data_construction}.

\subsection{AdvancedEdit Dataset}
\label{sec: AdvancedEdit Dataset}
We choose Pexels~\footnote{https://www.pexels.com/}, a high-quality real-world photographic image dataset, as our source data. Pexels is a popular platform offering a large collection of high-resolution, royalty-free images contributed by photographers around the world. The images in the Pexels dataset generally exhibit an average resolution of approximately 2K or higher. This dataset contains over 1 million photographic images and encompasses a broad range of subjects, covering various world scenes. Given that each image contains multiple objects and is associated with various instruction tasks, our dataset comprises a total of $2,536,674$ editing pairs. A more detailed analysis of the proposed dataset is provided in the appendix.

We define the image editing pairs with simple instructions as \textbf{SimpleEdit}, and image editing pairs with complex instructions as \textbf{AdvancedEdit}. We will discuss the influence of instructions with varying complexity in Section \ref{sec:ablation_datausage}. Additionally, we introduce the \textbf{AdvancedEdit-Eval}, comprising 300 curated image pairs, covering a range of tasks including removal, addition, and replacement. The evaluation dataset includes a variety of intricate and nuanced image editing scenarios, necessitating that the model possesses a certain level of understanding and reasoning capabilities.

\section{Method}
\noindent\textbf{Overview.}
\begin{figure*}[ht]    
    \centering    
    \setlength{\abovecaptionskip}{0cm}
    \resizebox{\textwidth}{!}{\includegraphics{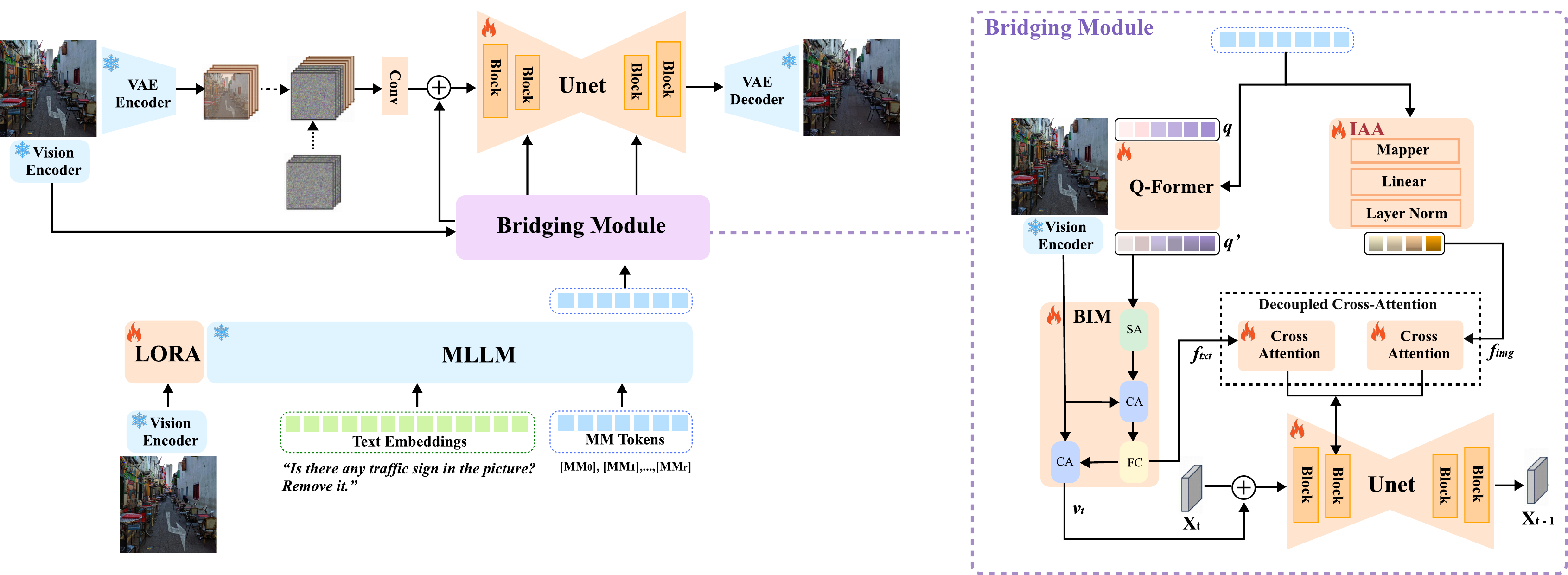}}    
    \caption{\textbf{The overall architecture of InsightEdit.} It mainly consists of three parts: (1) \textbf{Comprehension Module:} A comprehension module that leverages MLLM to perceive and comprehend the image editing task; (2) \textbf{Bridging Module:} A bridging module that better interacts and extracts both the textual and image features; (3) \textbf{Generation Module:} A generation module that receives editing guidance via diffusion model to generate the target image.}  
    \label{fig:method}    
\end{figure*}
The overall architecture of InsightEdit is depicted in Figure \ref{fig:method}. It mainly consists of a comprehension module, a bridging module, and a generation module. Specifically, the comprehension module leverages MLLM to comprehend the image editing task; the bridging module integrates both text and image features into the denoising process of the diffusion model; and the generation module receives editing guidance via the diffusion model to generate the target image.

\subsection{Comprehension Module}
\vspace{-1mm}
Multimodal large language model receives the original image and the editing instruction to comprehend the image editing task. In the comprehension module, we use LLaVA-7B~\cite{liu2024visual} as our vision-language foundation model. The feature of source image $\mathbf{I}_{src}$ is extracted by a vision encoder $E_\phi(\cdot)$ and the text instruction is tokenized as text embeddings $c$. Aligned by a fully connect layer, the original image features are sent into the LLaVA decoder~\cite{touvron2023llama} along with text embeddings.
The mentioned process is represented as:
\begin{equation}
\begin{aligned}
& v=\operatorname{FC}(E_\phi(\mathbf{I}_{src})), \\
& h=\operatorname{LLaVA}_{\omega}(v, c). \\
\end{aligned}
\end{equation}
Inspired by GILL ~\cite{koh2024generating}, we extend the vocabulary of the LLM by introducing \( r \) special [MM] tokens, where ``MM'' indicates multi-modality comprehension of text and image information via MLLM. 
These tokens are appended to the end of the instruction \( c \), and the model is trained by minimizing the negative log-likelihood of predicting [MM] tokens, conditioned on the previously generated tokens. The loss function can be represented below:
\begin{equation}
\begin{split}
    L_{\text{LLM}}(c) = - \sum_{i = 1}^{r} &\log p_{\{\omega \cup \mathbf{\theta}\}}([\text{MM}_i] \mid v, c,\\
        & [\text{MM}_1], \ldots, [\text{MM}_{i - 1}]) .
\end{split} 
\end{equation}
The majority of the LLM parameters are kept frozen, while the LoRA~\cite{hu2021lora} is employed to facilitate efficient training.

\begin{table*}[ht]
\setlength{\abovecaptionskip}{0cm}
\centering
\caption{\textbf{Quantitative comparison on AdvancedEdit-Eval.}}
\label{tab:comparison_advancededit}
\resizebox{0.7\textwidth}{!}{%
\begin{tabular}{lccccc}
\hline
\textbf{Methods}                     & \textbf{VIEScore↑} & \textbf{CLIPScore↑} & \textbf{PSNR↑} & \textbf{SSIM↑} & \textbf{LPIPS↓} \\ \hline
InstructPix2Pix             & 0.342      & 19.528      & 20.192    & 0.694 & 0.182  \\
MagicBrush                  & 0.352      & 19.751      & \underline{22.636}    & \textbf{0.743} & 0.132  \\
InstructDiffusion           & 0.318      & 19.390      & 18.025    & 0.624 & 0.223  \\
MGIE                        & 0.361      & 19.047      & 20.074    & 0.691 & 0.199  \\
SmartEdit-7B                & 0.682      & 20.114      & 20.115    & 0.651 & 0.131  \\ \hline
\textbf{InsightEdit}        & \underline{0.738}      & \underline{20.395}      & 21.267    & 0.675 & \underline{0.112}  \\
\textbf{InsightEdit with AdvancedEdit} & \textbf{0.831}      & \textbf{21.002}      & \textbf{22.871}    & \underline{0.716} & \textbf{0.071}  \\ \hline
\end{tabular}%
}
\end{table*}

\subsection{Bridging Module}
\vspace{-1mm}
The textual features provide high-level information while the image features provide more detailed condition. Therefore, in the bridging module, to comprehensively integrate both textual and image features as the conditions, we design a dual-stream condition alignment methodology. 



\noindent\textbf{Textual Branch via Q-Former \& BIM.} For the textual branch, similar to SmartEdit~\cite{huang2024smartedit}, we start from the \textit{text-aligned Q-Former}~\cite{li2023blip, carion2020end}  which has primarily aligned with the original clip text embedding space. Then, given the learnable tokens $q$
as inputs, we propose to further extract the reasoned textual information from the [MM] token hidden states $h$ via cross-attention:
\begin{equation}
     q' = Q_\beta(q, h), 
\end{equation}
where \( q' \) represents the Q-former outputs that mainly encode the text instruction information.

Additionally, following SmartEdit~\cite{huang2024smartedit}, we employ the BIM module that enables bidirectional information exchange between the input image and the MLLM output: 
\begin{equation}
    f_{txt}, v_{txt} = \operatorname{BIM}(E_\phi(\mathbf{I}_{src}),q'), 
\end{equation}
where $f_{txt}$ and $v_{txt}$ are the interacted features obtained from the BIM module. $f_{txt}$ is used as the final textual condition to UNet and $v_{txt}$ is the textual-aware vision features which is later added with U-Net inputs. 

\noindent\textbf{Image Branch via IAA.} For the image branch, we propose an \textit{Image Alignment Adapter (IAA)} module to activate the image information reasoned by MLLM, which consists of a mapper, a linear layer, and a layer normalization. 

In detail, the mapper is a Multi-Layer Perceptron (MLP) that transforms the [MM] token hidden states $h\in\mathbb{R}^{r\times4096}$ into a single embedding  $\operatorname{Mapper}(h) \in \mathbb{R}^{1\times768}$, which is then aligned with the target image features extracted by a CLIP vision encoder. Intuitively, the target image contains direct information about the editing goal, \textit{e.g.}, the expected background, therefore can provide explicit supervision. 
After that, the linear is further employed to expand the $\operatorname{Mapper}(h) \in \mathbb{R}^{1\times768}$ back to token sequences \( \mathbb{R}^{N\times768} \) ($N$ is set as $4$ by default) followed by a layer normalization. Formally, the mentioned process is represented as:
\begin{equation}
\begin{aligned}
\mathcal{L}_{\text{IAA}} &= \left\| \operatorname{CLIP}(\mathbf{I}_{\text{tar}}) - \operatorname{Mapper}(h) \right\|_2^2, \\
f_{img} &= \operatorname{IAA}(h),
\end{aligned}
\end{equation}
where \( \mathcal{L}_{\text{IAA}} \) is the supervision from the target image, and $f_{img}$ represents the output of the IAA, which will be sent to further cross-attention operation in U-Net.

Compared to textual features, the image features contain more detailed and comprehensive information, which can guide image editing more explicitly.

\subsection{Generation Module}
\vspace{-1mm}
In the process of target image generation, inspired by IP-adapter~\cite{ye2023ip}, we apply a decoupled cross-attention mechanism to integrate both textual and image features. An additional cross-attention layer is employed on top of the cross-attention layer in the original UNet block to interact with the image features. As the text cross-attention and image cross-attention are detached, we can also adjust the weight of the image condition in the inference stage:
\begin{equation}
\begin{split}
\mathbf{Z} &= \operatorname{Attention}(\mathbf{Q}, \mathbf{K_{txt}}, \mathbf{V_{txt}}) \\
           &\quad + \lambda \cdot \operatorname{Attention}(\mathbf{Q}, \mathbf{K_{img}}, \mathbf{V_{img}}),
\end{split}
\end{equation}
 where $\mathbf{K_{img}}$ and $\mathbf{V_{img}}$ are projected by $f_{img}$, $\mathbf{K_{txt}}$ and $\mathbf{V_{txt}}$ are projected by $f_{txt}$. 
 

For the diffusion model,  we concatenate the encoded image latent \( \mathcal{E}(\mathbf{I}_{src}) \) with the noisy latent \( z_t \). The process can be formulated as:

\begin{equation}
\begin{split}
\mathcal{L}_{\text{diff}} = & \mathbb{E}_{\mathcal{E}(y), \mathcal{E}(x), c_T, \epsilon \sim \mathcal{N}(0,1), t} \\
& \left[ \| \epsilon - \epsilon_\delta \left(t, \operatorname{concat}\left[z_t, \mathcal{E}(x)\right] + v_t, f_{txt}, f_{img} \right) \|_2^2 \right].
\end{split}
\end{equation}

\begin{table*}[ht]
\setlength{\abovecaptionskip}{0cm}
\centering
\caption{\textbf{Quantitative comparison on Reason-Edit.} All the methods we compared have been fine-tuned using the same training data as that used by SmartEdit. }
\label{tab:comparison_reasonedit} 
\resizebox{\textwidth}{!}{%
\begin{tabular}{l|ccccc|ccccc}  
\hline
\textbf{Methods} & \multicolumn{5}{c}{\textbf{Understanding Scenarios}} & \multicolumn{5}{c}{\textbf{Reasoning Scenarios}} \\ \cline{2-11}
 & \textbf{VIEScore↑} & \textbf{CLIPScore↑} & \textbf{PSNR↑} & \textbf{SSIM↑} & \textbf{LPIPS↓} & \textbf{VIEScore↑} & \textbf{CLIPScore↑} & \textbf{PSNR↑} & \textbf{SSIM↑} & \textbf{LPIPS↓} \\ \hline
InstructPix2Pix & 0.493 & 22.762 & 21.576 & 0.721 & 0.089 & 0.548 & 19.413 & 24.234 & 0.707 & 0.083 \\
MagicBrush & 0.422 & 22.620 & 18.120 & 0.680 & 0.143 & 0.530 & 19.755 & 22.101 & 0.694 & 0.113 \\
InstructDiffusion & 0.457 & 23.080 & 23.258 & 0.743 & \underline{0.067} & 0.670 & 19.523 & 21.453 & 0.666 & 0.117 \\
MGIE & 0.373 & 21.947 & 16.094 & 0.695 & 0.200 & 0.540 & 18.037 & 22.977 & 0.743 & 0.144 \\
SmartEdit-7B & 0.866 & 23.611 & 22.049 & 0.731 & 0.087 & 0.835 & 20.950 & 25.258 & 0.742 & 0.055 \\ \hline
\textbf{InsightEdit} & \underline{0.901} & \underline{23.734} & \underline{23.588} & \underline{0.749} & \underline{0.067} & \underline{0.893} & 20.867 & \underline{25.712} & \underline{0.747} & \underline{0.049} \\ 
\textbf{InsightEdit with AdvancedEdit} & \textbf{0.934} & \textbf{24.421} & \textbf{24.503} & \textbf{0.760} & \textbf{0.054} & \textbf{0.947} & \textbf{21.141} & \textbf{26.090} & \textbf{0.750} & \textbf{0.047} \\ \hline
\end{tabular}%
}
\end{table*}


\begin{figure*}[ht]    
    \setlength{\abovecaptionskip}{0cm}
    \centering    
    \includegraphics[width=0.9\linewidth]{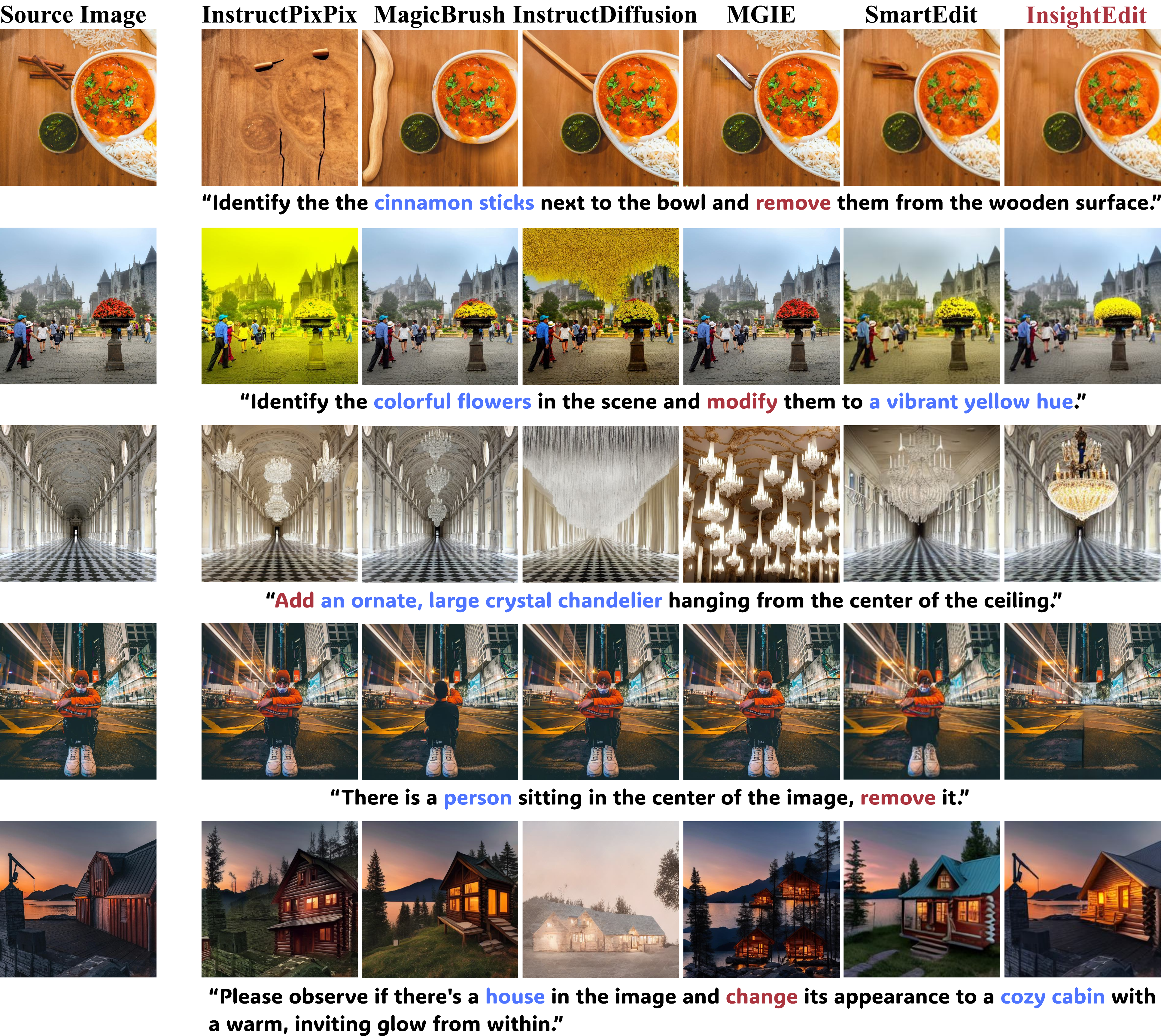}
    \caption{\textbf{Qualitative comparison on AdvancedEdit.} InsightEdit shows superior instruction following and background consistency capability.}  
    \label{fig:comparison_advanced_benchmark}    
    \vspace{-6mm}
\end{figure*}

\section{Results}

\subsection{Experiment Settings}
\vspace{-1mm}
\noindent\textbf{Implementation Details.} To train InsightEdit, we use CC12M~\cite{changpinyo2021conceptual}, InstructPix2Pix~\cite{brooks2023instructpix2pix}, MagicBrush~\cite{zhang2024magicbrush}, COCO~\cite{lin2014microsoft}, RefCOCO, GRefCOCO~\cite{kazemzadeh2014referitgame, yu2016modeling}, COCOStuff~\cite{caesar2018coco}, LISA~\cite{lai2023lisa, yang2023improved}, ReasonEdit~\cite{huang2024smartedit} and our proposed AdvancedEdit. Due to the limited resources, we only used $202,822$ editing pairs from AdvancedEdit for the experiments.

InsightEdit is trained in three stages, as detailed in the Appendix, and the training is conducted on a single machine with $8$ H100 GPUs. We adopt the AdamW optimizer~\cite{kingma2014adam} as the optimizer during three stages. In the first training stage, the learning rate is $2e^{-4}$ and weight decay is $0$. In the second stage and the third stage, the values of learning rate, weight decay, and warm-up ratio were set to $1e^{-5}$, $0$, and $0.001$, respectively. In the data construction pipeline, the MLLM we utilize is GPT-4o~\footnote{https://openai.com/}, while the LLM applied is Qwen2~\cite{yang2024qwen2}.

\noindent\textbf{Metrics.} For evaluating the background consistency, we use PSNR, SSIM, and LPIPS~\cite{hore2010image, zhang2018perceptual}. For the edited area, we calculate the CLIPScore~\cite{radford2021learning}  by comparing the edited object with its ground truth (GT) label. To more accurately evaluate editing effects and align with human preferences, we employ VIEScore~\cite{ku2023viescore}, which utilizes MLLM to evaluate editing performance from two perspectives: instruction-following and background consistency. \textbf{We emphasize that VIEScore might better align with human preferences, making it a more convincing measure~\cite{ku2023viescore}}.

\subsection{Comparison with State-of-the-art}
\vspace{-1mm}

\noindent\textbf{Results on AdvancedEdit-Eval.}
We compare our method with existing state-of-the-art instruction-based image methods, including instructPix2Pix, MagicBrush, InstructDiffusion, MGIE, and SmartEdit. First, we conduct comparative experiments on the AdvancedEdit-Eval as introduced in Section \ref{sec: AdvancedEdit Dataset}. The quantitative results presented in Table \ref{tab:comparison_advancededit} show that InsightEdit shows good performance across all five metrics. The experimental results in PSNR and LPIPS scores prove that InsightEdit excels in preserving non-edited regions, while the enhanced CLIPScore validates our method’s superior ability to follow instructions with greater accuracy. Specifically, in terms of VIEScore, both InsightEdit itself and InsightEdit trained on the AdvancedEdit dataset achieve state-of-the-art results, demonstrating that the model's generated outputs are preferred by humans over other methods, whether regarding background preservation or the ability to follow complex instructions.

Figure \ref{fig:comparison_advanced_benchmark} illustrates the qualititive results. As shown in the last column, our method trained on the AdvancedEdit dataset demonstrates a clear advantage over competing approaches. As can be seen, other methods struggle to maintain background consistency with the original image, often introducing artifacts or unintended alterations, while InsightEdit provides specific editing on the required regions. In the meanwhile, InsightEdit performs strong instruction following capabilities, having a great understanding of spatial, color, number, and detailed objects.

\noindent\textbf{Results on Reason-Edit.}
Additionally, we conducted comparative experiments on the Reason-Edit\cite{huang2024smartedit}, which is specifically designed to assess editing abilities with an emphasis on understanding and reasoning across various scenarios (\textit{i.e.} color, position, mirror, multiple-objects, reason, and size). The results in Table \ref{tab:comparison_reasonedit} show that 
InsightEdit outperforms other metrics, except for a slight CLIP score difference in reasoning tasks, where it slightly trails SmartEdit. Furthermore, when trained on our AdvancedEdit dataset, InsightEdit demonstrates improvements across all metrics, highlighting the model's ability to leverage the dataset to enhance both comprehension and reasoning capabilities, while also emphasizing the intrinsic value of the data used.

\subsection{Ablation Studies}
\vspace{-1mm}
\noindent\textbf{Ablation Studies on IAA.}
To validate the effectiveness of IAA, we conduct comparative experiments on InsightEdit with the AdvancedEdit dataset. Table \ref{tab:ablation_cam} shows the quantitive results, it can be seen that, with the IAA module, InsightEdit leads to notable improvements across various evaluation metrics. The qualitative results can be seen in Figure \ref{fig:ablation_cam}. In the first example, without the IAA module, the house is erroneously transformed into wildflowers, resulting in poor background consistency due to the absence of the red house. In contrast, with the IAA module, the red house is preserved accurately, maintaining consistency with the source image. In the second example, the marble table is edited poorly without the IAA module, while the addition of the IAA module leads to a more natural edit that aligns well with the instruction.

\begin{table}[t]
    \centering
    \scriptsize
 \setlength\tabcolsep{7.2pt} 
    \begin{tabular}{c|ccccc}
    \hline
    \textbf{IAA} & \textbf{PSNR↑} & \textbf{SSIM↑} & \textbf{LPIPS↓} & \textbf{CLIPScore↑} & \textbf{VIEScore↑} \\ \hline
                    & 22.348    & 0.692 & 0.095  & 20.652      & 7.307     \\
    \checkmark      & \textbf{22.871}    & \textbf{0.716} & \textbf{0.071}  & \textbf{21.002}      & \textbf{7.545}    \\ \hline
    \end{tabular}
    \setlength{\abovecaptionskip}{0cm} 
    \caption{\textbf{Ablation of IAA.} Our IAA module performs best.}
    \label{tab:ablation_cam}
\end{table}

\begin{figure}[t]
    \centering
    \setlength{\abovecaptionskip}{0cm}
    \includegraphics[width=0.9\linewidth]{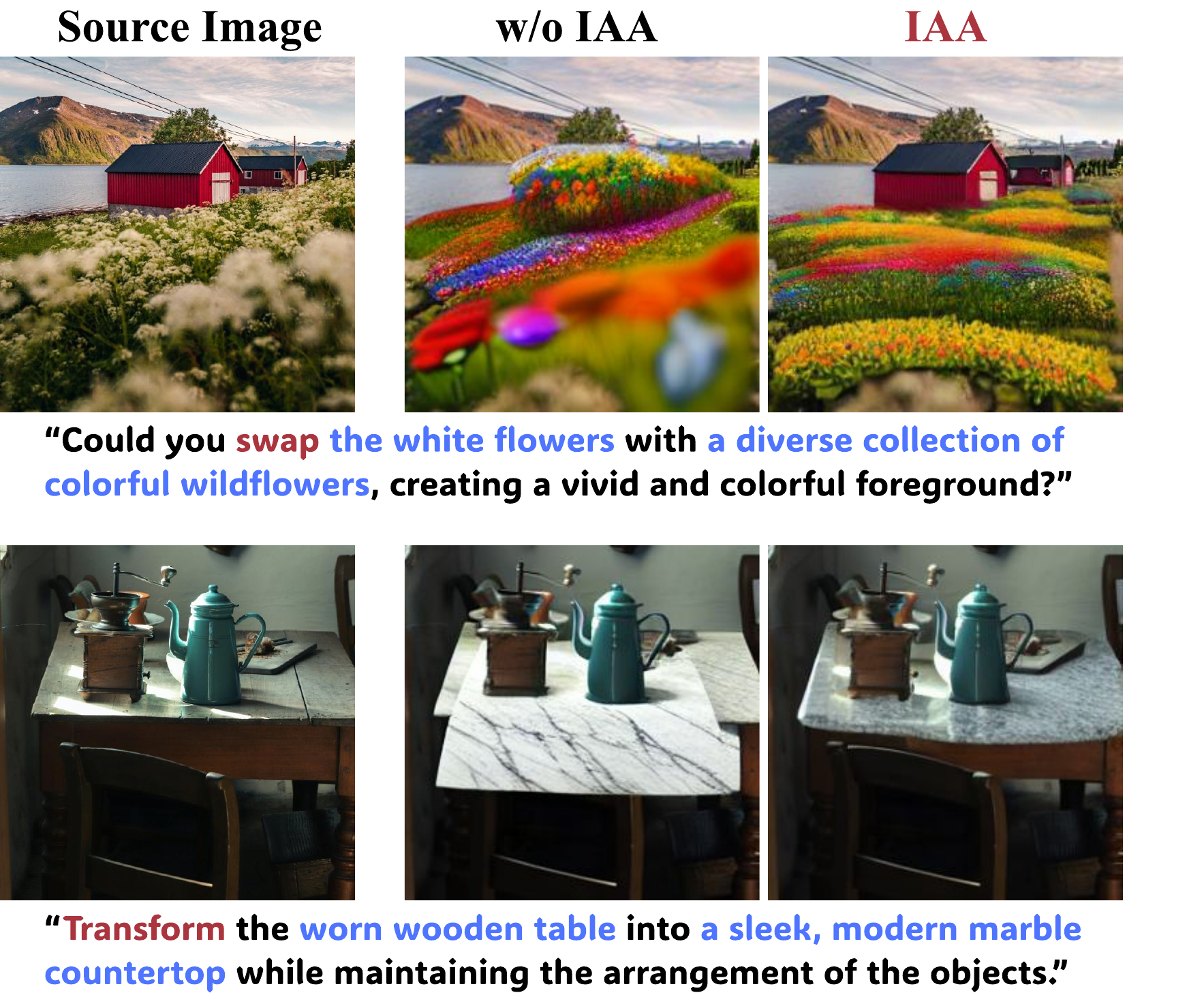} 
    \caption{Demonstration of the effectiveness of the IAA module.}
    \label{fig:ablation_cam}
\end{figure}

\noindent\textbf{Ablation Studies on Instruction Complexity.}
\label{sec:ablation_datausage}
We investigate the impact of instruction complexity and granularity on model performance by comparing two datasets introduced in Section \ref{sec:caption_mask_preparation}: SimpleEdit Dataset and AdvancedEdit Dataset. Table \ref{tab:ablation_data} demonstrates that incorporating the SimpleEdit Dataset alone enhances the model's performance, primarily due to the high-quality outputs generated by our mask-based image editing approach. However, the inclusion of the AdvancedEdit Dataset leads to substantial improvements, particularly in CLIPScore and VIEScore, highlighting the importance of increased instruction detail and the value of integrating complex instructions. The qualitative comparison results, as illustrated in Figure \ref{fig:ablation_data}, reveal that without any data augmentation, the method struggles to understand the instructions and produce effective edits in both two examples. When the SimpleEdit dataset is introduced, there is some improvement, though the edit quality remains suboptimal. However, with the use of the AdvancedEdit dataset, the method demonstrates the ability to accurately interpret editing instructions, resulting in vivid and precise edits. For instance, in the first example, the method produces finely detailed orange and black butterfly wings, while in the second example, it accurately identifies and removes a light source, maintaining harmony in the surrounding content. These results further show that advanced instruction can handle more complex image editing scenarios and generate more vivid editing results.

\begin{table}[t]
    \centering
    \scriptsize
    \setlength\tabcolsep{1.8pt}
    \begin{tabular}{cc|ccccc}
    \hline
    \textbf{SimpleEdit} & \textbf{AdvancedEdit} & \textbf{PSNR↑} & \textbf{SSIM↑} & \textbf{LPIPS↓} & \textbf{CLIPScore↑} & \textbf{VIEScore↑} \\ \hline
                    &             & 21.267    & 0.675 & 0.112  & 20.395      & 6.974     \\
     \checkmark    &             & 22.260    & 0.708 & 0.080  & 20.557      & 7.213     \\
                   & \checkmark  & \textbf{22.871}    & \textbf{0.716} & \textbf{0.071}  & \textbf{21.002}      & \textbf{7.545}     \\ \hline
    \end{tabular}
    \setlength{\abovecaptionskip}{0cm} 
    \caption{\textbf{Ablation of AdvancedEdit.} Our method with AdvancedEdit performs best.}
    \label{tab:ablation_data}
\end{table}
\vspace{-1mm}

\begin{figure}[t]
    \centering
    \setlength{\abovecaptionskip}{0cm}
    \includegraphics[width=1.0\linewidth]{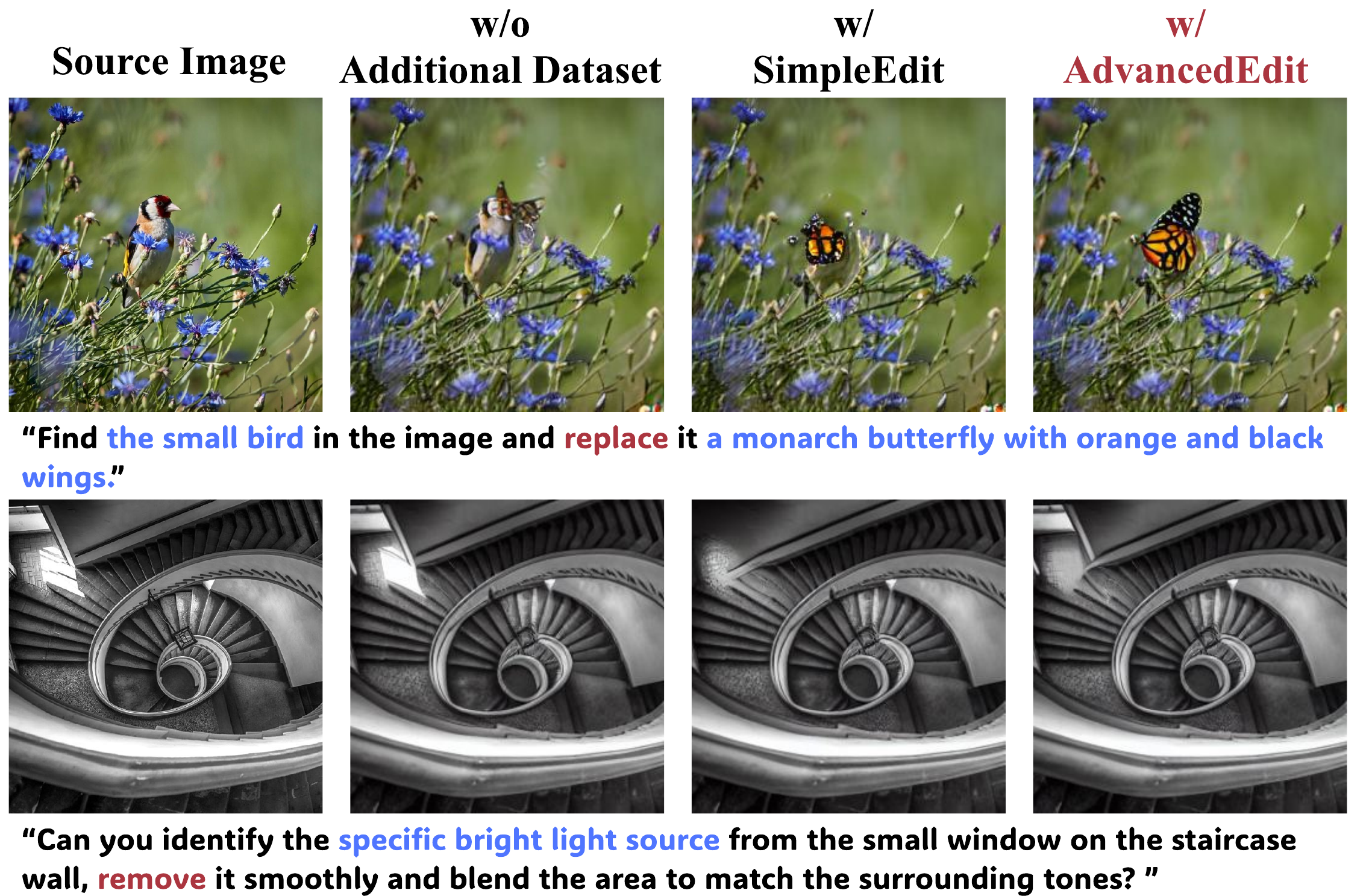} 
    \caption{Demonstration of the effectiveness of AdvancedEdit.}
    \label{fig:ablation_data}
\end{figure}

\section{Discussion and Conclusion}
In conclusion, we present InsightEdit, an end-to-end instruction-based image editing method, utilizing a two-stream bridging mechanism to integrate both the textual and visual features reasoned by the powerful MLLM into the diffusion model to guide the image editing process more precisely. We design an automatic data construction pipeline to generate a large-scale, high-quality, and high-fidelity image editing dataset, considering both image and instruction perspectives. We hope that our efforts will motivate more researchers in the future. \\
\noindent\textbf{Limitation and Future Work.} Several challenges remain to be addressed, including the potential for incorporating more advanced MLLM for an improved understanding of instructions. Additionally, upgrading to more powerful diffusion models could potentially enhance the editing quality. 


{
    \small
    \bibliographystyle{ieeenat_fullname}
    \bibliography{main}

\begin{thebibliography}{44}
\providecommand{\natexlab}[1]{#1}
\providecommand{\url}[1]{\texttt{#1}}
\expandafter\ifx\csname urlstyle\endcsname\relax
  \providecommand{\doi}[1]{doi: #1}\else
  \providecommand{\doi}{doi: \begingroup \urlstyle{rm}\Url}\fi

\bibitem[Avrahami et~al.(2023)Avrahami, Fried, and Lischinski]{avrahami2023blended}
Omri Avrahami, Ohad Fried, and Dani Lischinski.
\newblock Blended latent diffusion.
\newblock \emph{ACM transactions on graphics (TOG)}, 42\penalty0 (4):\penalty0 1--11, 2023.

\bibitem[Brooks et~al.(2023)Brooks, Holynski, and Efros]{brooks2023instructpix2pix}
Tim Brooks, Aleksander Holynski, and Alexei~A Efros.
\newblock Instructpix2pix: Learning to follow image editing instructions.
\newblock In \emph{Proceedings of the IEEE/CVF Conference on Computer Vision and Pattern Recognition}, pages 18392--18402, 2023.

\bibitem[Caesar et~al.(2018)Caesar, Uijlings, and Ferrari]{caesar2018coco}
Holger Caesar, Jasper Uijlings, and Vittorio Ferrari.
\newblock Coco-stuff: Thing and stuff classes in context.
\newblock In \emph{Proceedings of the IEEE conference on computer vision and pattern recognition}, pages 1209--1218, 2018.

\bibitem[Cao et~al.(2023)Cao, Wang, Qi, Shan, Qie, and Zheng]{cao2023masactrl}
Mingdeng Cao, Xintao Wang, Zhongang Qi, Ying Shan, Xiaohu Qie, and Yinqiang Zheng.
\newblock Masactrl: Tuning-free mutual self-attention control for consistent image synthesis and editing.
\newblock In \emph{Proceedings of the IEEE/CVF International Conference on Computer Vision}, pages 22560--22570, 2023.

\bibitem[Carion et~al.(2020)Carion, Massa, Synnaeve, Usunier, Kirillov, and Zagoruyko]{carion2020end}
Nicolas Carion, Francisco Massa, Gabriel Synnaeve, Nicolas Usunier, Alexander Kirillov, and Sergey Zagoruyko.
\newblock End-to-end object detection with transformers.
\newblock In \emph{European conference on computer vision}, pages 213--229. Springer, 2020.

\bibitem[Chakrabarty et~al.(2023)Chakrabarty, Singh, Saakyan, and Muresan]{chakrabarty2023learning}
Tuhin Chakrabarty, Kanishk Singh, Arkadiy Saakyan, and Smaranda Muresan.
\newblock Learning to follow object-centric image editing instructions faithfully.
\newblock \emph{arXiv preprint arXiv:2310.19145}, 2023.

\bibitem[Changpinyo et~al.(2021)Changpinyo, Sharma, Ding, and Soricut]{changpinyo2021conceptual}
Soravit Changpinyo, Piyush Sharma, Nan Ding, and Radu Soricut.
\newblock Conceptual 12m: Pushing web-scale image-text pre-training to recognize long-tail visual concepts.
\newblock In \emph{Proceedings of the IEEE/CVF conference on computer vision and pattern recognition}, pages 3558--3568, 2021.

\bibitem[Fu et~al.(2023)Fu, Hu, Du, Wang, Yang, and Gan]{fu2023guiding}
Tsu-Jui Fu, Wenze Hu, Xianzhi Du, William~Yang Wang, Yinfei Yang, and Zhe Gan.
\newblock Guiding instruction-based image editing via multimodal large language models.
\newblock \emph{arXiv preprint arXiv:2309.17102}, 2023.

\bibitem[Geng et~al.(2024)Geng, Yang, Hang, Li, Gu, Zhang, Bao, Zhang, Li, Hu, et~al.]{geng2024instructdiffusion}
Zigang Geng, Binxin Yang, Tiankai Hang, Chen Li, Shuyang Gu, Ting Zhang, Jianmin Bao, Zheng Zhang, Houqiang Li, Han Hu, et~al.
\newblock Instructdiffusion: A generalist modeling interface for vision tasks.
\newblock In \emph{Proceedings of the IEEE/CVF Conference on Computer Vision and Pattern Recognition}, pages 12709--12720, 2024.

\bibitem[Hertz et~al.(2022)Hertz, Mokady, Tenenbaum, Aberman, Pritch, and Cohen-Or]{hertz2022prompt}
Amir Hertz, Ron Mokady, Jay Tenenbaum, Kfir Aberman, Yael Pritch, and Daniel Cohen-Or.
\newblock Prompt-to-prompt image editing with cross attention control.
\newblock \emph{arXiv preprint arXiv:2208.01626}, 2022.

\bibitem[Hore and Ziou(2010)]{hore2010image}
Alain Hore and Djemel Ziou.
\newblock Image quality metrics: Psnr vs. ssim.
\newblock In \emph{2010 20th international conference on pattern recognition}, pages 2366--2369. IEEE, 2010.

\bibitem[Hu et~al.(2021)Hu, Shen, Wallis, Allen-Zhu, Li, Wang, Wang, and Chen]{hu2021lora}
Edward~J Hu, Yelong Shen, Phillip Wallis, Zeyuan Allen-Zhu, Yuanzhi Li, Shean Wang, Lu Wang, and Weizhu Chen.
\newblock Lora: Low-rank adaptation of large language models.
\newblock \emph{arXiv preprint arXiv:2106.09685}, 2021.

\bibitem[Huang et~al.(2024{\natexlab{a}})Huang, Huang, Liu, Yan, Lv, Liu, Xiong, Zhang, Chen, and Cao]{huang2024diffusion}
Yi Huang, Jiancheng Huang, Yifan Liu, Mingfu Yan, Jiaxi Lv, Jianzhuang Liu, Wei Xiong, He Zhang, Shifeng Chen, and Liangliang Cao.
\newblock Diffusion model-based image editing: A survey.
\newblock \emph{arXiv preprint arXiv:2402.17525}, 2024{\natexlab{a}}.

\bibitem[Huang et~al.(2024{\natexlab{b}})Huang, Xie, Wang, Yuan, Cun, Ge, Zhou, Dong, Huang, Zhang, et~al.]{huang2024smartedit}
Yuzhou Huang, Liangbin Xie, Xintao Wang, Ziyang Yuan, Xiaodong Cun, Yixiao Ge, Jiantao Zhou, Chao Dong, Rui Huang, Ruimao Zhang, et~al.
\newblock Smartedit: Exploring complex instruction-based image editing with multimodal large language models.
\newblock In \emph{Proceedings of the IEEE/CVF Conference on Computer Vision and Pattern Recognition}, pages 8362--8371, 2024{\natexlab{b}}.

\bibitem[Hui et~al.(2024)Hui, Yang, Zhao, Shi, Wang, Wang, Zhou, and Xie]{hui2024hq}
Mude Hui, Siwei Yang, Bingchen Zhao, Yichun Shi, Heng Wang, Peng Wang, Yuyin Zhou, and Cihang Xie.
\newblock Hq-edit: A high-quality dataset for instruction-based image editing.
\newblock \emph{arXiv preprint arXiv:2404.09990}, 2024.

\bibitem[Ju et~al.(2024)Ju, Liu, Wang, Bian, Shan, and Xu]{ju2024brushnet}
Xuan Ju, Xian Liu, Xintao Wang, Yuxuan Bian, Ying Shan, and Qiang Xu.
\newblock Brushnet: A plug-and-play image inpainting model with decomposed dual-branch diffusion.
\newblock \emph{arXiv preprint arXiv:2403.06976}, 2024.

\bibitem[Kazemzadeh et~al.(2014)Kazemzadeh, Ordonez, Matten, and Berg]{kazemzadeh2014referitgame}
Sahar Kazemzadeh, Vicente Ordonez, Mark Matten, and Tamara Berg.
\newblock Referitgame: Referring to objects in photographs of natural scenes.
\newblock In \emph{Proceedings of the 2014 conference on empirical methods in natural language processing (EMNLP)}, pages 787--798, 2014.

\bibitem[Kingma(2014)]{kingma2014adam}
Diederik~P Kingma.
\newblock Adam: A method for stochastic optimization.
\newblock \emph{arXiv preprint arXiv:1412.6980}, 2014.

\bibitem[Koh et~al.(2024)Koh, Fried, and Salakhutdinov]{koh2024generating}
Jing~Yu Koh, Daniel Fried, and Russ~R Salakhutdinov.
\newblock Generating images with multimodal language models.
\newblock \emph{Advances in Neural Information Processing Systems}, 36, 2024.

\bibitem[Ku et~al.(2023)Ku, Jiang, Wei, Yue, and Chen]{ku2023viescore}
Max Ku, Dongfu Jiang, Cong Wei, Xiang Yue, and Wenhu Chen.
\newblock Viescore: Towards explainable metrics for conditional image synthesis evaluation.
\newblock \emph{arXiv preprint arXiv:2312.14867}, 2023.

\bibitem[Lai et~al.(2023)Lai, Tian, Chen, Li, Yuan, Liu, and Jia]{lai2023lisa}
Xin Lai, Zhuotao Tian, Yukang Chen, Yanwei Li, Yuhui Yuan, Shu Liu, and Jiaya Jia.
\newblock Lisa: Reasoning segmentation via large language model.
\newblock \emph{arXiv preprint arXiv:2308.00692}, 2023.

\bibitem[Li et~al.(2023)Li, Li, Savarese, and Hoi]{li2023blip}
Junnan Li, Dongxu Li, Silvio Savarese, and Steven Hoi.
\newblock Blip-2: Bootstrapping language-image pre-training with frozen image encoders and large language models.
\newblock In \emph{International conference on machine learning}, pages 19730--19742. PMLR, 2023.

\bibitem[Lin et~al.(2014)Lin, Maire, Belongie, Hays, Perona, Ramanan, Doll{\'a}r, and Zitnick]{lin2014microsoft}
Tsung-Yi Lin, Michael Maire, Serge Belongie, James Hays, Pietro Perona, Deva Ramanan, Piotr Doll{\'a}r, and C~Lawrence Zitnick.
\newblock Microsoft coco: Common objects in context.
\newblock In \emph{Computer Vision--ECCV 2014: 13th European Conference, Zurich, Switzerland, September 6-12, 2014, Proceedings, Part V 13}, pages 740--755. Springer, 2014.

\bibitem[Liu et~al.(2024)Liu, Li, Wu, and Lee]{liu2024visual}
Haotian Liu, Chunyuan Li, Qingyang Wu, and Yong~Jae Lee.
\newblock Visual instruction tuning.
\newblock \emph{Advances in neural information processing systems}, 36, 2024.

\bibitem[Mann et~al.(2020)Mann, Ryder, Subbiah, Kaplan, Dhariwal, Neelakantan, Shyam, Sastry, Askell, Agarwal, et~al.]{mann2020language}
Ben Mann, N Ryder, M Subbiah, J Kaplan, P Dhariwal, A Neelakantan, P Shyam, G Sastry, A Askell, S Agarwal, et~al.
\newblock Language models are few-shot learners.
\newblock \emph{arXiv preprint arXiv:2005.14165}, 1, 2020.

\bibitem[Radford et~al.(2021)Radford, Kim, Hallacy, Ramesh, Goh, Agarwal, Sastry, Askell, Mishkin, Clark, et~al.]{radford2021learning}
Alec Radford, Jong~Wook Kim, Chris Hallacy, Aditya Ramesh, Gabriel Goh, Sandhini Agarwal, Girish Sastry, Amanda Askell, Pamela Mishkin, Jack Clark, et~al.
\newblock Learning transferable visual models from natural language supervision.
\newblock In \emph{International conference on machine learning}, pages 8748--8763. PMLR, 2021.

\bibitem[Ren et~al.(2024)Ren, Liu, Zeng, Lin, Li, Cao, Chen, Huang, Chen, Yan, Zeng, Zhang, Li, Yang, Li, Jiang, and Zhang]{ren2024grounded}
Tianhe Ren, Shilong Liu, Ailing Zeng, Jing Lin, Kunchang Li, He Cao, Jiayu Chen, Xinyu Huang, Yukang Chen, Feng Yan, Zhaoyang Zeng, Hao Zhang, Feng Li, Jie Yang, Hongyang Li, Qing Jiang, and Lei Zhang.
\newblock Grounded sam: Assembling open-world models for diverse visual tasks, 2024.

\bibitem[Rombach et~al.(2022)Rombach, Blattmann, Lorenz, Esser, and Ommer]{rombach2022high}
Robin Rombach, Andreas Blattmann, Dominik Lorenz, Patrick Esser, and Bj{\"o}rn Ommer.
\newblock High-resolution image synthesis with latent diffusion models.
\newblock In \emph{Proceedings of the IEEE/CVF conference on computer vision and pattern recognition}, pages 10684--10695, 2022.

\bibitem[Ronneberger et~al.(2015)Ronneberger, Fischer, and Brox]{ronneberger2015u}
Olaf Ronneberger, Philipp Fischer, and Thomas Brox.
\newblock U-net: Convolutional networks for biomedical image segmentation.
\newblock In \emph{Medical image computing and computer-assisted intervention--MICCAI 2015: 18th international conference, Munich, Germany, October 5-9, 2015, proceedings, part III 18}, pages 234--241. Springer, 2015.

\bibitem[Touvron et~al.(2023)Touvron, Lavril, Izacard, Martinet, Lachaux, Lacroix, Rozi{\`e}re, Goyal, Hambro, Azhar, et~al.]{touvron2023llama}
Hugo Touvron, Thibaut Lavril, Gautier Izacard, Xavier Martinet, Marie-Anne Lachaux, Timoth{\'e}e Lacroix, Baptiste Rozi{\`e}re, Naman Goyal, Eric Hambro, Faisal Azhar, et~al.
\newblock Llama: Open and efficient foundation language models.
\newblock \emph{arXiv preprint arXiv:2302.13971}, 2023.

\bibitem[Wang et~al.(2023)Wang, Saharia, Montgomery, Pont-Tuset, Noy, Pellegrini, Onoe, Laszlo, Fleet, Soricut, et~al.]{wang2023imagen}
Su Wang, Chitwan Saharia, Ceslee Montgomery, Jordi Pont-Tuset, Shai Noy, Stefano Pellegrini, Yasumasa Onoe, Sarah Laszlo, David~J Fleet, Radu Soricut, et~al.
\newblock Imagen editor and editbench: Advancing and evaluating text-guided image inpainting.
\newblock In \emph{Proceedings of the IEEE/CVF conference on computer vision and pattern recognition}, pages 18359--18369, 2023.

\bibitem[Xie et~al.(2023{\natexlab{a}})Xie, Zhang, Lin, Hinz, and Zhang]{xie2023smartbrush}
Shaoan Xie, Zhifei Zhang, Zhe Lin, Tobias Hinz, and Kun Zhang.
\newblock Smartbrush: Text and shape guided object inpainting with diffusion model.
\newblock In \emph{Proceedings of the IEEE/CVF Conference on Computer Vision and Pattern Recognition}, pages 22428--22437, 2023{\natexlab{a}}.

\bibitem[Xie et~al.(2023{\natexlab{b}})Xie, Zhao, Xiao, Chan, Li, Xu, Zhang, and Hou]{xie2023dreaminpainter}
Shaoan Xie, Yang Zhao, Zhisheng Xiao, Kelvin~CK Chan, Yandong Li, Yanwu Xu, Kun Zhang, and Tingbo Hou.
\newblock Dreaminpainter: Text-guided subject-driven image inpainting with diffusion models.
\newblock \emph{arXiv preprint arXiv:2312.03771}, 2023{\natexlab{b}}.

\bibitem[Yang et~al.(2024{\natexlab{a}})Yang, Yang, Hui, Zheng, Yu, Zhou, Li, Li, Liu, Huang, et~al.]{yang2024qwen2}
An Yang, Baosong Yang, Binyuan Hui, Bo Zheng, Bowen Yu, Chang Zhou, Chengpeng Li, Chengyuan Li, Dayiheng Liu, Fei Huang, et~al.
\newblock Qwen2 technical report.
\newblock \emph{arXiv preprint arXiv:2407.10671}, 2024{\natexlab{a}}.

\bibitem[Yang et~al.(2024{\natexlab{b}})Yang, Zeng, Liu, Li, Xu, Zhang, and Yan]{yang2024editworld}
Ling Yang, Bohan Zeng, Jiaming Liu, Hong Li, Minghao Xu, Wentao Zhang, and Shuicheng Yan.
\newblock Editworld: Simulating world dynamics for instruction-following image editing.
\newblock \emph{arXiv preprint arXiv:2405.14785}, 2024{\natexlab{b}}.

\bibitem[Yang et~al.(2023)Yang, Qu, Lai, Tian, Peng, Liu, and Jia]{yang2023improved}
Senqiao Yang, Tianyuan Qu, Xin Lai, Zhuotao Tian, Bohao Peng, Shu Liu, and Jiaya Jia.
\newblock An improved baseline for reasoning segmentation with large language model.
\newblock \emph{arXiv preprint arXiv:2312.17240}, 2023.

\bibitem[Ye et~al.(2023)Ye, Zhang, Liu, Han, and Yang]{ye2023ip}
Hu Ye, Jun Zhang, Sibo Liu, Xiao Han, and Wei Yang.
\newblock Ip-adapter: Text compatible image prompt adapter for text-to-image diffusion models.
\newblock \emph{arXiv preprint arXiv:2308.06721}, 2023.

\bibitem[Yu et~al.(2016)Yu, Poirson, Yang, Berg, and Berg]{yu2016modeling}
Licheng Yu, Patrick Poirson, Shan Yang, Alexander~C Berg, and Tamara~L Berg.
\newblock Modeling context in referring expressions.
\newblock In \emph{Computer Vision--ECCV 2016: 14th European Conference, Amsterdam, The Netherlands, October 11-14, 2016, Proceedings, Part II 14}, pages 69--85. Springer, 2016.

\bibitem[Zhang et~al.(2024{\natexlab{a}})Zhang, Mo, Chen, Sun, and Su]{zhang2024magicbrush}
Kai Zhang, Lingbo Mo, Wenhu Chen, Huan Sun, and Yu Su.
\newblock Magicbrush: A manually annotated dataset for instruction-guided image editing.
\newblock \emph{Advances in Neural Information Processing Systems}, 36, 2024{\natexlab{a}}.

\bibitem[Zhang et~al.(2023)Zhang, Rao, and Agrawala]{zhang2023adding}
Lvmin Zhang, Anyi Rao, and Maneesh Agrawala.
\newblock Adding conditional control to text-to-image diffusion models.
\newblock In \emph{Proceedings of the IEEE/CVF International Conference on Computer Vision}, pages 3836--3847, 2023.

\bibitem[Zhang et~al.(2018)Zhang, Isola, Efros, Shechtman, and Wang]{zhang2018perceptual}
Richard Zhang, Phillip Isola, Alexei~A Efros, Eli Shechtman, and Oliver Wang.
\newblock The unreasonable effectiveness of deep features as a perceptual metric.
\newblock In \emph{CVPR}, 2018.

\bibitem[Zhang et~al.(2024{\natexlab{b}})Zhang, Yang, Feng, Qin, Chen, Yu, Chen, Wang, Savarese, Ermon, et~al.]{zhang2024hive}
Shu Zhang, Xinyi Yang, Yihao Feng, Can Qin, Chia-Chih Chen, Ning Yu, Zeyuan Chen, Huan Wang, Silvio Savarese, Stefano Ermon, et~al.
\newblock Hive: Harnessing human feedback for instructional visual editing.
\newblock In \emph{Proceedings of the IEEE/CVF Conference on Computer Vision and Pattern Recognition}, pages 9026--9036, 2024{\natexlab{b}}.

\bibitem[Zhao et~al.(2024)Zhao, Ma, Chen, Si, Wu, An, Yu, Zhang, Li, and Chang]{zhao2024ultraedit}
Haozhe Zhao, Xiaojian Ma, Liang Chen, Shuzheng Si, Rujie Wu, Kaikai An, Peiyu Yu, Minjia Zhang, Qing Li, and Baobao Chang.
\newblock Ultraedit: Instruction-based fine-grained image editing at scale.
\newblock \emph{arXiv preprint arXiv:2407.05282}, 2024.

\bibitem[Zhuang et~al.(2023)Zhuang, Zeng, Liu, Yuan, and Chen]{zhuang2023task}
Junhao Zhuang, Yanhong Zeng, Wenran Liu, Chun Yuan, and Kai Chen.
\newblock A task is worth one word: Learning with task prompts for high-quality versatile image inpainting.
\newblock \emph{arXiv preprint arXiv:2312.03594}, 2023.

\end{thebibliography}
}


\end{document}